\documentclass[runningheads]{llncs}
\usepackage{float}
\usepackage{longtable}
\usepackage{graphicx}
\usepackage{subfiles}
\usepackage{url}

\usepackage{adjustbox}
\newcommand*\rot{\rotatebox{90}}
\begin{document}
\title{Back to Basics: A Sanity Check on Modern Time Series Classification Algorithms}
%
%
%
\author{Bhaskar Dhariyal \and
Thach Le Nguyen \and
Georgiana Ifrim}
\institute{School of Computer Science, University College Dublin, Ireland\\
\email{bhaskar.dhariyal@ucdconnect.ie}, \email{\{thach.lenguyen, georgiana.ifrim\}@ucd.ie}}
%
%
\maketitle              
\begin{abstract}
The state-of-the-art in time series classification has come a long way, from the 1NN-DTW algorithm to the ROCKET family of classifiers. However, in the current fast-paced development of new classifiers, taking a step back and performing simple baseline checks is essential. These checks are often overlooked, as researchers are focused on establishing new state-of-the-art results, developing scalable algorithms, and making models explainable. Nevertheless, there are many datasets that look like time series at first glance, but classic algorithms such as tabular methods with no time ordering may perform better on such problems. For example, for spectroscopy datasets, tabular methods tend to significantly outperform recent time series methods. 
In this study, we compare the performance of tabular models using classic machine learning approaches (e.g., Ridge, LDA, RandomForest) with the ROCKET family of classifiers (e.g., Rocket, MiniRocket, MultiRocket). Tabular models are simple and very efficient, while the ROCKET family of classifiers are more complex and have state-of-the-art accuracy and efficiency among recent time series classifiers. We find that tabular models outperform the ROCKET family of classifiers on approximately 19\% of univariate and 28\% of multivariate datasets in the UCR/UEA benchmark and achieve accuracy within 10 percentage points on about 50\% of datasets. Our results suggest that it is important to consider simple tabular models as baselines when developing time series classifiers. These models are very fast, can be as effective as more complex methods and may be easier to understand and deploy.

\end{abstract}
\section{Introduction}


Time series classification is a challenging task that has attracted significant research interest recently. The ever-evolving computational capabilities and abundant applications and use cases have led to the development of a wide range of time series classification methods, from simple distance-based methods (1-NN-DTW \cite{sakoe1978dynamic}) to complex deep learning models (Inception Time \cite{ismail2020inceptiontime}).

Most of the research in time series classification is focused on establishing state-of-the-art results, developing scalable algorithms, and making models explainable. However, in this quest, it is often possible to forget the first principle of research, which is to compare with existing simpler methods. 

Historically, there have been many instances where traditional models have outperformed deep learning methods on some tasks. For example, a recent study \cite{Zeng2022AreTE} showed that linear models can be more effective than deep learning networks for forecasting. Similarly, the work of \cite{frizzarin2023classification} showed that linear models can outperform other complex models for classification tasks in spectroscopy data. However, there is less empirical  work investigating the performance of classic tabular models on time series classification tasks.

In this study, we take a step back from the pursuit of providing yet another state-of-the-art method and perform some simple sanity checks, which are often missed. We compare the performance of tabular models with the ROCKET \cite{dempster2019rocket,tan2022multirocket,dempster2021minirocket} family of classifiers, which are currently considered state-of-the-art for time series classification.
In this paper, the main contributions are:

\begin{itemize}
    \item We empirically compared tabular and time series methods on the established UCR/UEA benchmarks for univariate and multivariate time series classification.
    \item We analysed the accuracy-time tradeoffs for all the methods on both benchmarks and found that on about 50\% of datasets in both benchmarks, the tabular methods perform within 10 percentage points accuracy of state-of-the-art time series classification methods, while being two orders of magnitude faster.
    \item We discussed the performance of tabular versus time series methods for different data and problem types and the potential implications for how the very popular UCR/UEA benchmarks are formed and used by the community. In particular, if tabular methods significantly outperform time series methods on some problem types, we raise the question of whether these datasets should be included in a time series benchmark.
\end{itemize}


\section{Related Work}

\textbf{The UCR and UEA benchmarks.}
\textbf{Univariate Time Series Classification (UTSC).} State-of-the-art UTS classifiers are classifiers that have been shown to be the most accurate methods on the UCR/UEA benchmark. The most notable ones are ROCKET \cite{dempster2019rocket} and its variants (MiniROCKET, MultiROCKET and HYDRA \cite{Dempster2023}), due to their high accuracy and efficiency. These classifiers follow a two-step approach: transforming the time series into tabular features and classifying these transformations using linear models such as logistic regression.  While deep learning methods (e.g., FCN, ResNet, InceptionTime \cite{ismail2020inceptiontime}) or ensembles (e.g., HIVE-COTE \cite{Middlehurst2021}, TDE \cite{uea75490}) are also as accurate, they often demand significantly more computing resources (time, CPU, GPU, etc.). Other notable classifiers include symbolic-classifiers such as WEASEL \cite{schafer2023weasel} and MrSQM \cite{sqm} and shapelet-classifiers such as RDST \cite{rdst}. 
The UCR/UEA time series archive is a public collection of time series datasets that has been used extensively as the unified benchmark by researchers in this area. The archive is the result of a massive collaborative effort lead by research groups from the University California Riverside (UCR) and the University of East Anglia (UEA), hence the name of the benchmark. Starting with 85 univariate datasets in 2015, the archive was expanded to 128 datasets in 2018. The expansion also introduced a classification benchmark for multivariate time series which includes 30 datasets. The dedicated website\footnote{\url{http://www.timeseriesclassification.com}} for the archive contains not only the downloadable datasets but also pointers to code, publications, and other information that can be useful to any interested party. Without a doubt, the archive is a major resource that pushes forward research in TSC. However, while extremely useful for providing an overview and comparing against existing work, it potentially creates a pitfall where new works only focus on "beating the benchmark" and neglect what makes a classifier useful in real-life applications.

\textbf{Multivariate Time Series Classification (MTSC).}
In general, it can be said that the MTSC literature is less developed when compared to UTSC. The benchmark for MTSC was introduced later with fewer datasets. Most state-of-the-art MTSC methods are UTSC methods that are adapted for MTS data. The most straightforward approach is to learn from each channel independently (e.g., HIVE-COTE, WEASEL-MUSE \cite{schafer2017multivariate}). On the other hand, some classifiers actually utilize channel dependency, and thus are called bespoke MTS classifiers. For example, the multivariate variants of ROCKET (and MiniROCKET, MultiROCKET) replace the 1D kernels with 2D kernels to produce multi-channel dependent features (see \cite{dempster2021minirocket,tan2022multirocket} for details).   

\textbf{Tabular Methods.}
Classic machine learning models such as Random Forest, Logistic Regression, Linear Regression, seem to have been largely ignored in recent time series literature.  Such methods often assume independence between values at different time points and thus are deemed unsuitable for time series data. The work in \cite{bagnall2012transformation} employs tabular models, however, the models are trained on transformed data after applying techniques such as PCA, Spectral approaches and auto-correlation. Nonetheless, outside of the time series literature, these methods are still favourable choices in some communities. In particular, the work of \cite{frizzarin2023classification,FRIZZARIN2021104442} investigated several approaches for modelling milk spectroscopy data and found that tabular methods significantly outperformed time series methods. While these datasets are not inherently time series data, spectroscopy data have been part of the UCR/UEA benchmark since its inception and have been widely accepted by the community as time series data. This finding suggests that not all datasets in the benchmark are suitable for time series methods. We further investigate this issue in the next sections.




\section{Background}



A \textbf{time series} is a sequence of numbers representing some measurements over time. For example, a time series could represent a person's heartbeat variation on a 30-minute morning run. Each value in a time series usually has significance with respect to the previous and next values.

A typical mathematical representation of time series is $T: \{x_{0}, x_{1}, x_{2}, \ldots x_{n}\}$ where $x\in \Re$ and $n$ is the length of the time series. When we assign a discrete label to the time series, we can perform time series classification. We discuss two types of time series tasks in this paper, i.e., univariate time series classification (UTSC) and multivariate time series classification (MTSC). 
In univariate time series classification, data is recorded from a single source, meaning only one observed variable exists. On the other hand, multivariate time series classification involves recording data from multiple sources, resulting in the presence of multiple observed variables. A mathematical representation of multivariate time series can take the form:

$T: \{<x_0^0, x_1^0, \ldots x_n^0><x_0^1, x_1^1, \ldots x_n^1> \ldots  <x_0^{m-1}, x_1^{m-1}, \ldots x_n^{m-1}> \}$

where $m$ is the number of channels. If the time series is univariate, $m=1$. It is common in some applications to convert multivariate time series to univariate time series by concatenating all the channels into a single univariate time series. After this transformation, univariate classifiers can be trained with this data.

\textbf{Tabular data} is the most ubiquitous data type. It is a data structure that organizes data into rows and columns. Each row represents a single record, and each column represents a single attribute of that record. It has  no concept of temporality. This means that the previous value has no impact on the current value. A time series can be considered a tabular vector and used as input to a tabular method, e.g., linear regression.

\section{Experiments}

\subsection{Datasets}

The UEA/UCR \cite{dau2019ucr} benchmark datasets are mostly used in the empirical evaluation and comparison of various algorithms. Since the benchmark contains both univariate and multivariate datasets, it is popular for testing new algorithms on. Table \ref{tab:ddic_mtsc} and \ref{tab:ddic_univariate} in the appendix provide the data dictionary for both types of datasets. As it is common in recent time series literature, we run experiments on 109 univariate datasets and 25 equal-length multivariate datasets. We make our code available on github\footnote{\url{https://github.com/mlgig/TabularModelsforTSC}}.

\subsection{Univariate Time Series Classification}
\label{ref:region_def}

Before comparing tabular versus time-series models, we compared a few popular methods within each group separately. 

\begin{figure}[ht!]
    \centering
    \includegraphics[width=0.8\textwidth]{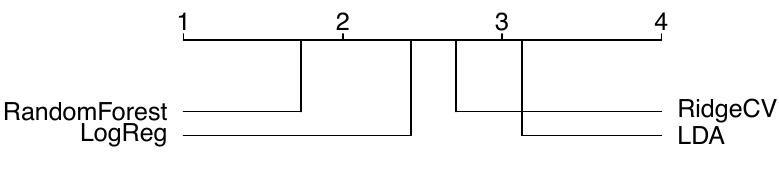}
    \caption{Accuracy comparison of tabular methods on UTSC datasets.}
    \label{fig:tab_method_utsc}
\end{figure}

\textbf{Tabular Methods Results.} For tabular methods we select three linear methods known for their efficiency and effectiveness in real-world applications \cite{frizzarin2023classification}, as well as Random Forest to have an effective non-linear classifier. We run these methods using the sklearn implementation\footnote{\url{https://scikit-learn.org/stable/supervised_learning.html}} with default parameters. Later in the paper we also discuss parameter tuning and its impact on accuracy and runtime.
In Figure \ref{fig:tab_method_utsc}, we compare the accuracy of four tabular models on univariate datasets: Random Forest, Logistic Regression, Ridge Regression (RidgeCV) and Latent Dirichlet Analysis (LDA). 
The critical difference diagram \cite{demvsar2006statistical} captures the average accuracy rank over all the datasets. The accuracy gain is evaluated using a Wilcoxon signed-rank test with Holm correction and visualised with the critical difference (CD) diagram with significance value ($\alpha$) = 0.05. The figure illustrates Random Forest significantly outperforms the other three models and Logistic Regression outperforms the other linear models 
Table \ref{tab:utsc_tab_time_acc} illustrates the mean accuracy and total training and test computation time in minutes. The tabular results correspond to the tabular CD diagram, where Random Forest is the best classifier.


\begin{table}[ht!]
    \caption{Mean accuracy and total computation time taken by tabular models on UTSC datasets.}
    \centering
    \begin{tabular}{c|cc}
    \hline
    & Mean Accuracy & Total Time (minutes) \\
    \hline
    RandomForest & 0.74 & 0.886\\
    LogReg & 0.69 & 0.31 \\
    RidgeCV & 0.67 & 0.09 \\
    LDA & 0.63 & 0.09 \\
    \hline
\end{tabular}

    \label{tab:utsc_tab_time_acc}
\end{table}

\begin{figure}[ht!]
    \centering
    \includegraphics[width=0.7\textwidth]{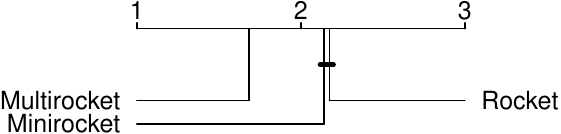}
    \caption{Accuracy comparison of time-series methods on UTSC datasets.}
    \label{fig:ts_method_utsc}
\end{figure}

\begin{table}[ht!]
    \centering
    \caption{Mean accuracy and total computation time taken by time-series models on UTSC datasets.}
    \begin{tabular}{c|cc}
    \hline
    & Mean Accuracy & Total Time (minutes) \\
    \hline
    Minirocket & 0.86 & 34.56 \\
    Multirocket & 0.86 & 73.46 \\
    Rocket & 0.85 & 158.76 \\
    \hline
    \end{tabular}
    \label{tab:utsc_time_acc_time}
\end{table}

\begin{figure}
    \centering
    \includegraphics[width=0.8\textwidth]{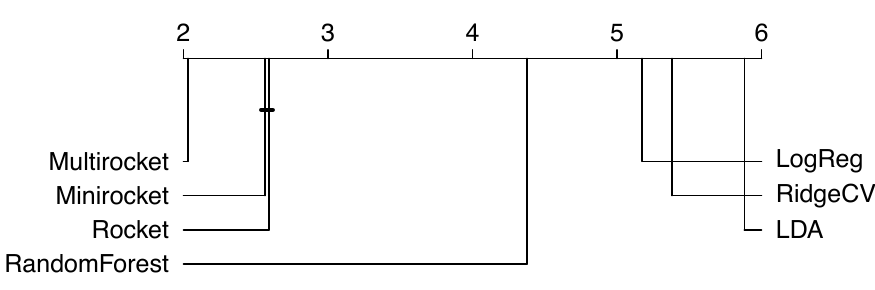}
    \caption{Accuracy comparison of tabular and time series models on UTSC datasets.}
    \label{fig:tab_ts_utsc}
\end{figure}

\textbf{Time Series Methods Results.} Similarly, in Figure \ref{fig:ts_method_utsc} and Table \ref{tab:utsc_time_acc_time}, we compare the accuracy of three time series classification models: Multirocket, MiniRocket, and Rocket. 
We use the implementation in the aeon-toolkit library\footnote{\url{https://www.aeon-toolkit.org/en/latest/api_reference/classification.html}} with default parameters.
From the critical difference diagram (Figure \ref{fig:ts_method_utsc}) we note that MultiRocket is significantly more accurate  than MiniRocket and Rocket.

\subsubsection{Time Series Methods vs Tabular Methods.}

In  Figure \ref{fig:tab_ts_utsc}, we compare the accuracy of time-series and tabular models. We can see that the time-series models have a higher mean accuracy rank than the tabular models. Multirocket is significantly more accurate than all other models, and Random Forest is the closest tabular model to the time-series models.

\subsubsection{Detailed Analysis.}
Figure \ref{fig:tab_ts_utsc} provides a summary overview of the performance of classifiers using their average accuracy ranking across the datasets analysed. Average behaviour with respect to accuracy or rank is a common and useful summary to get an overview of the performance of multiple classifiers over multiple datasets.
However, it is crucial to examine the performance of models at a finer level to understand the difference in behaviour between tabular and time-series models. 

\begin{figure}[!htbp]
    \centering
    \includegraphics[width=\textwidth]{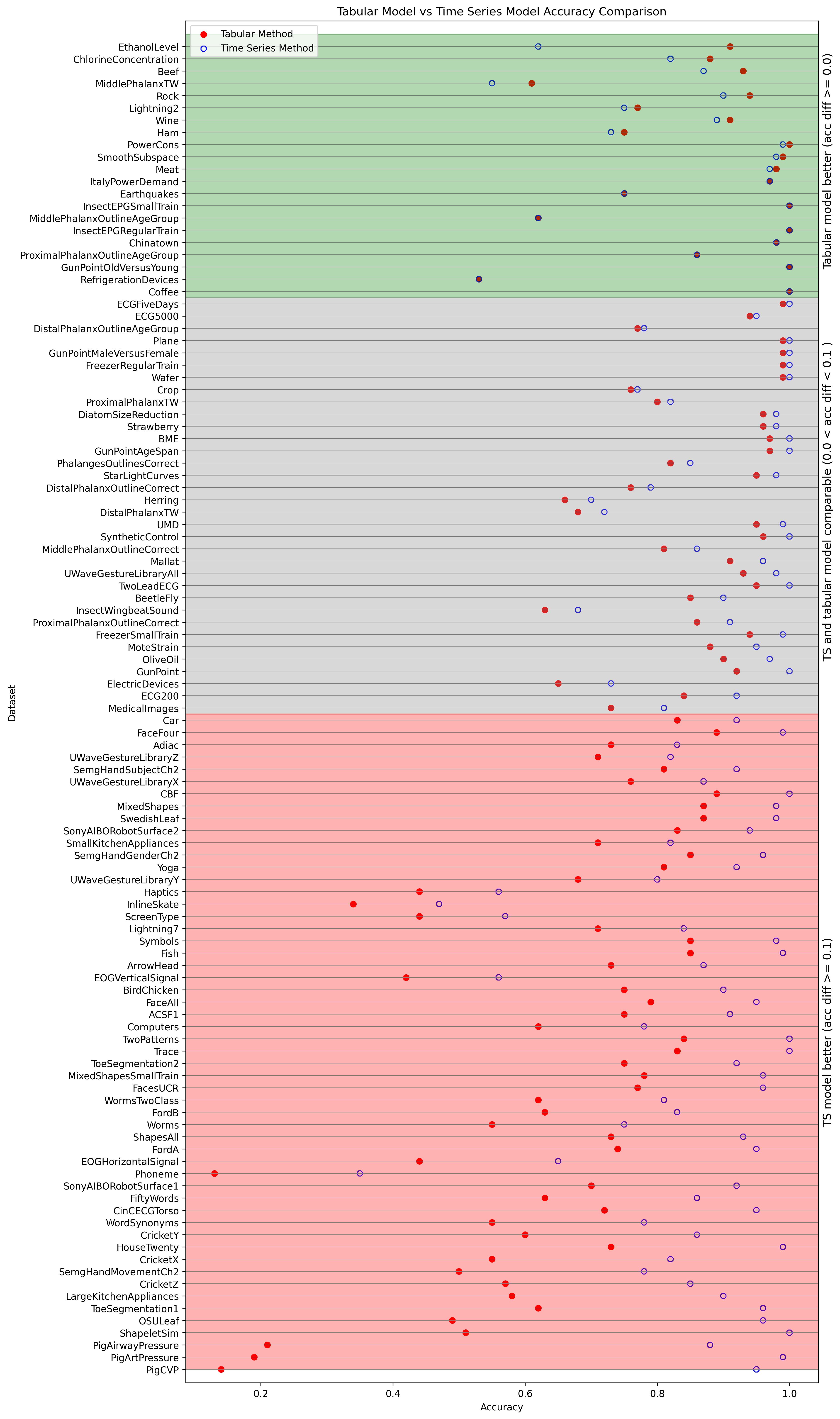}
    \caption{Accuracy comparison of the best time series model with the best tabular model on univariate time series datasets. Red circles represent the tabular models, and blue circles represent the time series models. Each marker shows the maximum accuracy achieved by the tabular models versus the time series models.}
    \label{fig:dataset_utsc_mtsc}
\end{figure}

In Figure \ref{fig:dataset_utsc_mtsc}, we illustrate the accuracy of tabular and time series models on each dataset, focusing on comparing the best-performing tabular with the best-performing time series model. The plot is divided into three distinct regions: green, grey, and red.

\begin{itemize}

\item The green region illustrates the datasets where the tabular models outperform the time series models or where both models achieve the same accuracy. 

\item The grey region represents datasets where the two models have performance within a fixed threshold. It is crucial to consider the accuracy-time trade-off in this region when deciding the better model. Datasets in this region are highlighted when the difference between the best-performing time-series model and the best-performing tabular model ranges from 1 to 9 percentage points.

\item The red region represents the datasets where time series models outperform tabular models. The time series models in these datasets are at least 10 percentage points better than tabular models.

\end{itemize}

For the UEA benchmark, surprisingly, 19.2\% of the datasets performed better with tabular models (green region), 31.1\% performed within 10 percentage points with both tabular and time series models (grey region), and 49.5\% performed better than 10 percentage points with time series models (red region).

The above numbers imply that on about 19\% of the benchmark, there are only weak temporal patterns, and tabular methods that disregard time ordering are very competitive when compared with time series methods. 
As a result, for many of those datasets in the green and grey region, using a complex time series model would be like using a sledgehammer to crack a nut. We of course acknowledge that time series methods work very well for the datasets in the red region, but these account for slightly less than half of the benchmark. We also acknowledge that the Rocket algorithms have been tested outside of this benchmark with good results in many real time series applications \cite{DBLP:journals/datamine/SinghBNHMOWCI23,DBLP:journals/datamine/DhariyalNI23,DBLP:journals/corr/BagnallBLL16,DBLP:journals/datamine/RuizFLMB21}. The question remains though: should we include the datasets in the green and grey areas into a time series benchmark at all, given that tabular methods have similar accuracy to the best time series methods on those datasets.

\subsubsection{Computation Time Analysis.} 
Traditionally, tabular models are known for their computational speed. This is also evident from Tables \ref{tab:utsc_tab_time_acc} and \ref{tab:utsc_time_acc_time}, which show that tabular models are an order of magnitude faster than time series models. Figure \ref{fig:dataset_utsc_mtsc} illustrates the various regions for accuracy, but it is worth highlighting that tabular models in the green and grey regions are faster and almost as accurate, or even more accurate than time series methods.



\begin{figure}[!htbp]
    \centering
    \includegraphics[width=\textwidth]{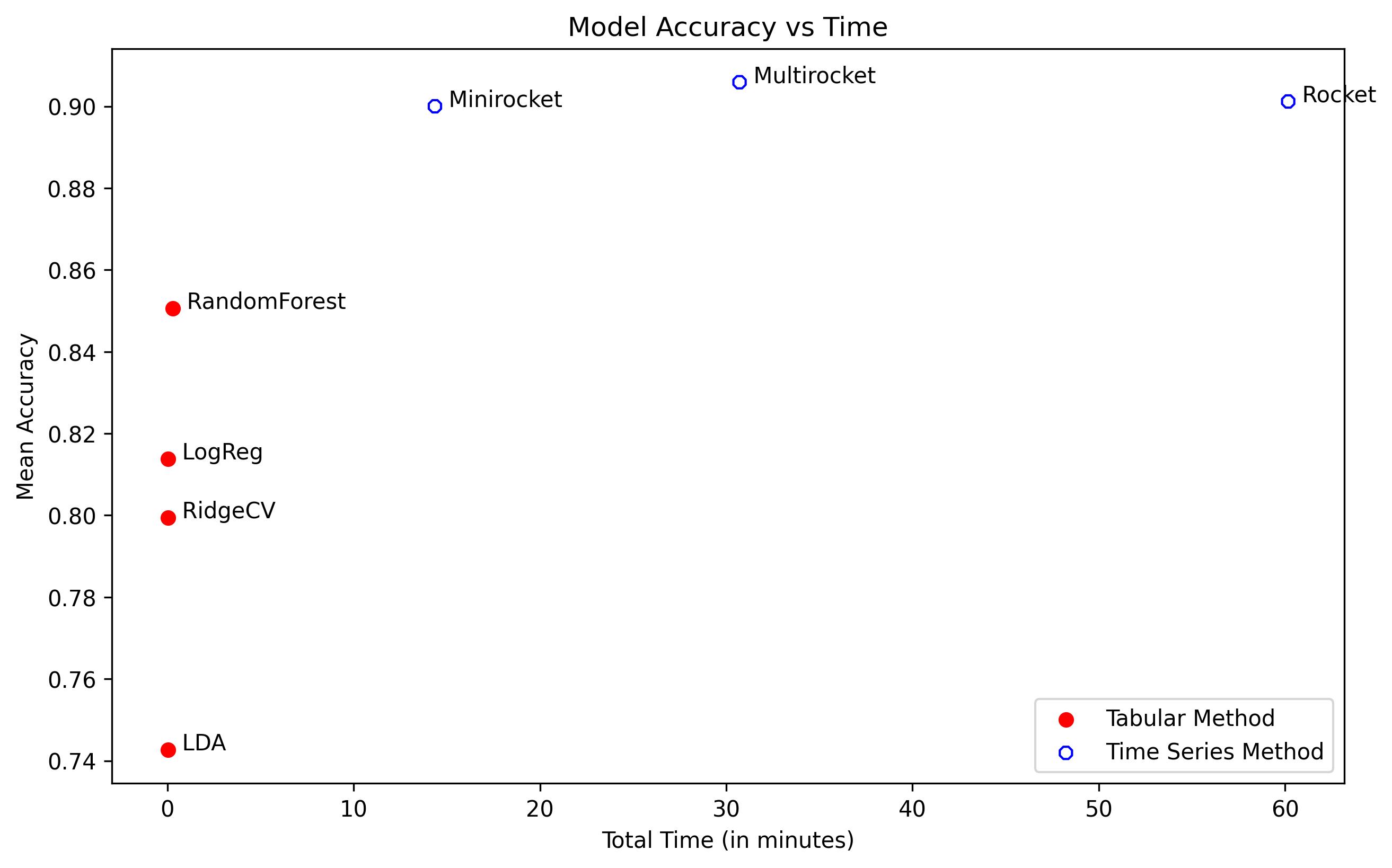}
    \caption{Accuracy-Time tradeoff for datasets in the \textbf{grey region} shown in Figure \ref{fig:dataset_utsc_mtsc}. We observe a mean accuracy difference of about 5 percentage points, but at least an order of magnitude difference in computation time, between tabular and time series methods.}
    \label{fig:acc_time_tradeoff_uts}
\end{figure}

Figure \ref{fig:acc_time_tradeoff_uts} shows the tradeoff between the mean accuracy and total computation time for the various time-series and tabular models in grey region datasets. Multirocket and Random Forest are the most accurate models among time series and tabular models, respectively. The difference in accuracy between Multirocket and Random Forest is approximately 5 percentage points. However, Multirocket takes an average of 30 minutes longer to train.

\subsubsection{Domain-wise Analysis.}
Table \ref{tab:utsc_domain}  shows the mean accuracy of different classifiers on datasets from various domains (as annotated by the meta-data in UCR/UEA). The benchmark is highly dominated by three domains: Image, Sensor, and Motion. About 63\%  of the benchmark comprises these three domains out of a total of 13 domains in the benchmark. 

\begin{table}[H]
    \centering
    \caption{Mean accuracy of classifiers by problem types on UCR univariate datasets.}
\begin{tabular}{c|cccc|ccc}
\hline
& \multicolumn{4}{c|}{Tabular Models} & \multicolumn{3}{c}{Time Series Models}\\
\hline
Domain (\#datasets)   &   RidgeCV &   LDA &   LogReg &   RandomForest &   Rocket &   Minirocket &   Multirocket \\
\hline
 Image(31) &      0.66 &  0.62 &     \textbf{0.71} &           0.75 &     0.85 &         0.85 &          \textbf{0.85} \\
 Sensor(20) &      0.73 &  0.69 &     0.72 &           \textbf{0.76} &     0.86 &         0.86 &          \textbf{0.87} \\
 Motion(17) &      0.58 &  0.46 &     0.58 &           \textbf{0.70}  &     0.84 &         0.84 &          \textbf{0.85} \\
 Device(8) &      0.48 &  0.44 &     0.48 &           0.62 &     0.76 &         0.74 &          \textbf{0.77} \\
 Simulated(8) &      0.78 &  0.82 &     0.81 &           \textbf{0.88} &     \textbf{0.99} &         0.98 &          \textbf{0.99} \\
 Spectro(8) &      0.86 &  \textbf{0.90}  &     0.86 &           0.82 &     0.84 &         0.86 &          \textbf{0.86} \\
 ECG(4) &      \textbf{0.92} &  0.84 &     0.92 &           0.82 &     0.97 &         0.97 &          \textbf{0.97} \\
 Spectrum(4) &      \textbf{0.75} &  0.67 &     0.74 &           0.67 &     0.83 &         0.82 &          \textbf{0.88} \\
 Hemodynamics(3) &      0.05 &  \textbf{0.16} &     0.12 &           0.13 &     0.66 &         \textbf{0.94} &          0.81 \\
 EOG(2) &      0.3  &  0.28 &     0.37 &           \textbf{0.43} &     0.59 &         0.57 &          \textbf{0.60}  \\
 EPG(2) &      0.82 & \textbf{1.00}   &    \textbf{1.00}   &          1.00   &     0.99 &        1.00   &         1.00   \\
 Power(1) &      0.98 &  0.73 &     0.99 &          \textbf{1.00}   &     0.92 &         \textbf{0.99} &          0.98 \\
 Traffic(1) &      \textbf{0.98} &  0.95 &     0.98 &           0.98 &     0.98 &         0.98 &          \textbf{0.98} \\
\hline
\end{tabular}

    \label{tab:utsc_domain}
\end{table}

As expected, with regard to average accuracy in a specific domain, as also shown in Figure \ref{fig:tab_ts_utsc}, time series models performed better than tabular models in most of the domains. However, we note that the tabular models performed especially well in the Spectro domain. This could be because the Spectro domain does not have strong temporal features. Also, as we have seen in Figure \ref{fig:dataset_utsc_mtsc}, average behaviour can be misleading and we need to look at the accuracy on individual datasets to get a good idea of accuracy behaviour across the entire benchmark or specific domains.

\subsection{Multivariate Time Series Classification}

In addition to our analysis of univariate time series datasets, we also conducted an analysis on multivariate time series datasets. The UEA/UCR benchmark dataset we utilized for this analysis consisted of 26 datasets. However, to ensure consistency and comparability among the models, we narrowed down our focus to the 25 datasets that all models could run on. We filtered the datasets based on equal length, and one dataset (Pen Digits) was removed due to  Minirocket, which cannot run on datasets with lengths less than 8.

\textbf{Data Preprocessing:} Unlike univariate time series, which have data from a single channel, multivariate time series data have multiple channels. To convert this data into a format that a tabular model can process, we first standardize each channel's data and then concatenate the data across all channels. 

\begin{figure}[h!]
    \centering
    \includegraphics[width=0.8\textwidth]{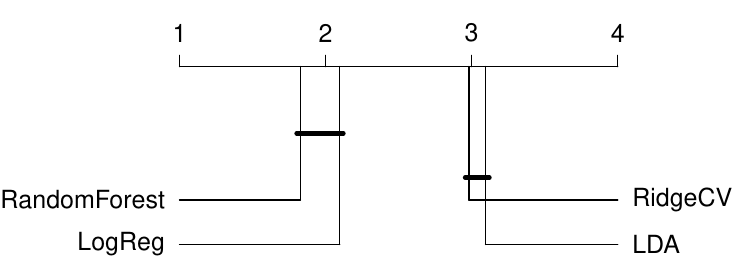}
    \caption{Accuracy comparison of tabular methods on MTSC datasets.}
    \label{fig:rmtsc_tab}
\end{figure}

\textbf{Tabular Methods Results.} After preprocessing the data, we followed a similar approach to our univariate analysis. We selected the same tabular models: Random Forest, LDA, Logistic Regression, and RidgeCV. The critical difference diagram (Figure \ref{fig:rmtsc_tab}) illustrates that Random Forest performed significantly better than the other three models, and Logistic Regression outperformed the other two linear models.

Table \ref{tab:mtsc_acc_time} shows the total time taken by tabular models and their corresponding mean accuracy. The table corroborates the results of the critical difference diagram, which showed that Random Forest is the most accurate tabular model, closely followed by LogisticRegression and RidgeCV. RidgeCV is also the most time-efficient method.

\begin{table}[h!]
    \centering
        \caption{Mean accuracy and total computation time taken by tabular models on MTSC
datasets.}
    \begin{tabular}{c|cc}
    \hline
    & Mean Accuracy & Total Time (minutes) \\
    \hline
    RandomForest & 0.61 & 6.40 \\
    LogisticRegression & 0.59 & 6.20 \\
    RidgeCV & 0.56 & 5.27 \\
    LDA & 0.52 & 6.70 \\
    \hline
    \end{tabular}

    \label{tab:mtsc_acc_time}
\end{table}

\textbf{Time Series Methods Results.} Similar to the tabular methods, we ran the multivariate time series methods, namely Minirocket, Multirocket, and Rocket, on the MTSC datasets. Since the implemented algorithm works well with multivariate time series, there was no need to preprocess the data in this case.

\begin{figure}[h!]
    \centering
    \includegraphics[width=0.7\textwidth]{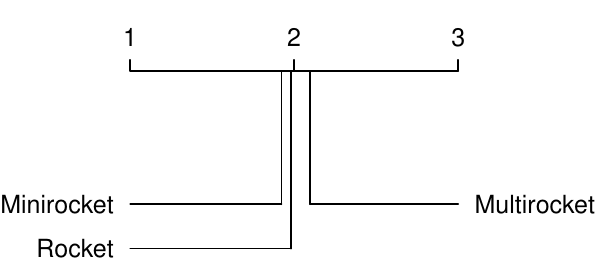}
    \caption{Accuracy comparison of time-series methods on MTSC datasets.}
    \label{fig:rmtsc_time}
\end{figure}

Figure \ref{fig:rmtsc_time} and Table \ref{tab:rmtsc_time_acc} illustrate the performance of time series methods on the benchmark datasets. Both the figure and table show that Minirocket outperforms the other two classifiers. Additionally, Minirocket is also the fastest method among the three methods.

\begin{table}[h!]
    \centering
        \caption{Mean accuracy and total computation time taken by time series models on MTSC
datasets.}
    \begin{tabular}{c|cc}
    \hline
    & Mean Accuracy & Total Time (minutes) \\    
    \hline
    Minirocket & 0.71 & 49.33 \\
    Multirocket & 0.70 & 67.10 \\
    Rocket & 0.70 & 129.05 \\
    \hline
    \end{tabular}

    \label{tab:rmtsc_time_acc}
\end{table}

\subsubsection{Time Series Methods vs Tabular Methods.} 
\begin{figure}[h!]
    \centering
    \includegraphics[width=0.8\textwidth]{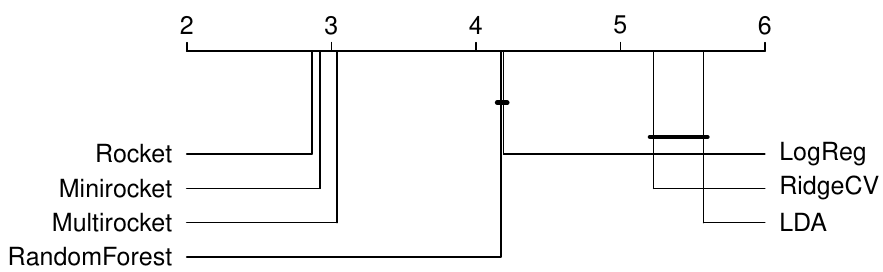}
    \caption{Accuracy comparison time-series and tabular methods on MTSC datasets.}
    \label{fig:rmtsc_time_tab}
\end{figure}

Finally, we compared tabular and time series models, as shown in Figure \ref{fig:rmtsc_time_tab}. As expected, the time series models outperformed the tabular models in terms of average accuracy. However, we conducted a more detailed analysis to investigate the reasons for this difference. We  discuss our findings below.

\begin{figure}[t!]
    \centering
    \includegraphics[width=\textwidth]{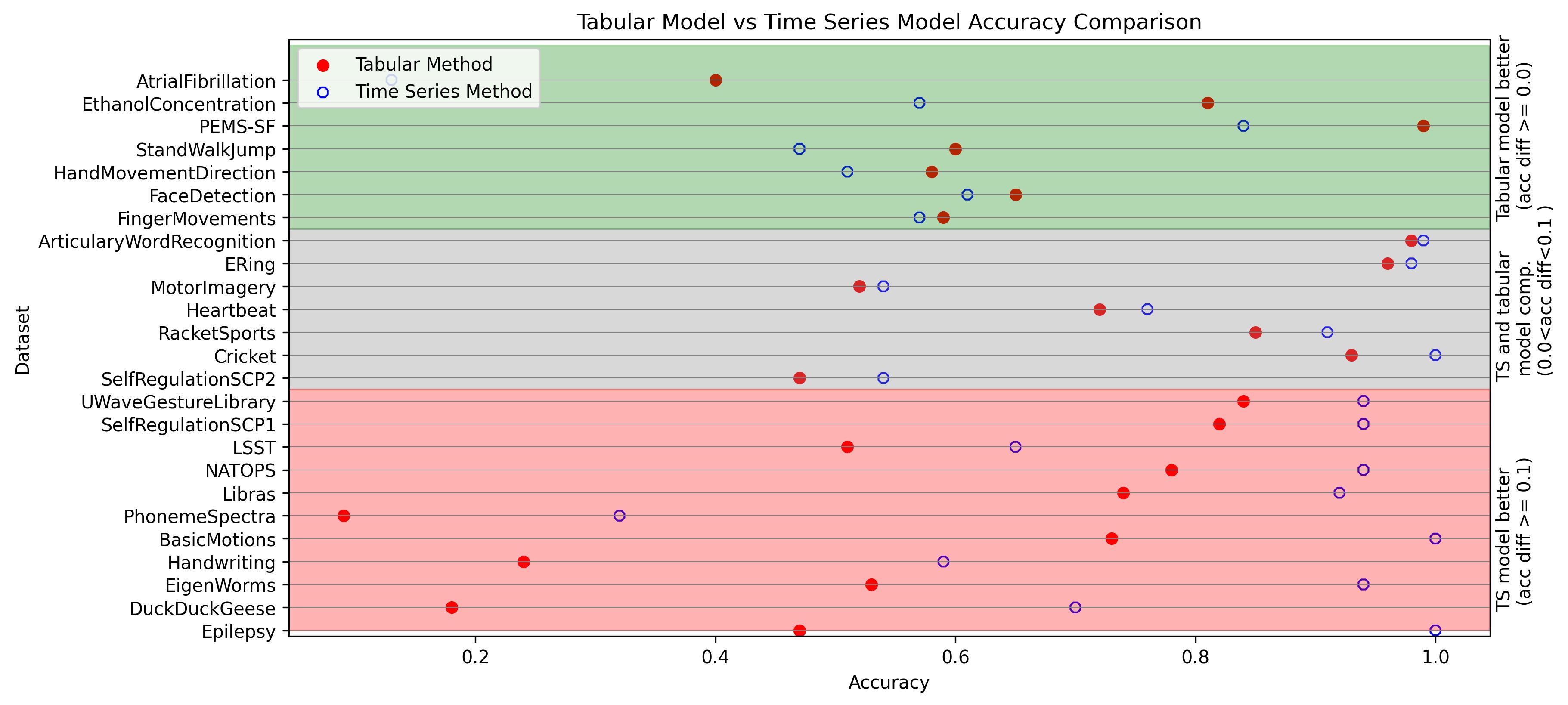}
    \caption{Accuracy comparison of time series models with tabular models on multivariate time series datasets. Red circles represent tabular models, and blue circles represent time series models. Each marker shows the maximum accuracy achieved by the tabular models versus the time series models. Detailed results are provided with the code.}
    \label{fig:lm_ts_mtsc}
\end{figure}

Figure \ref{fig:lm_ts_mtsc} shows the difference in performance between the best-performing tabular model and the best-performing time series model. The performance of each model is highlighted in a different region, as defined above in Section \ref{ref:region_def}. Approximately 28 percent of the datasets are represented in each green and grey region (56 percent total), indicating that the tabular model performs better or within 10 percentage points in these cases. Another 44 percent of the datasets fall within the red region, indicating that the time series models outperform the tabular models in those instances. 

\begin{figure}[th]
    \centering
    \includegraphics[width=\textwidth]{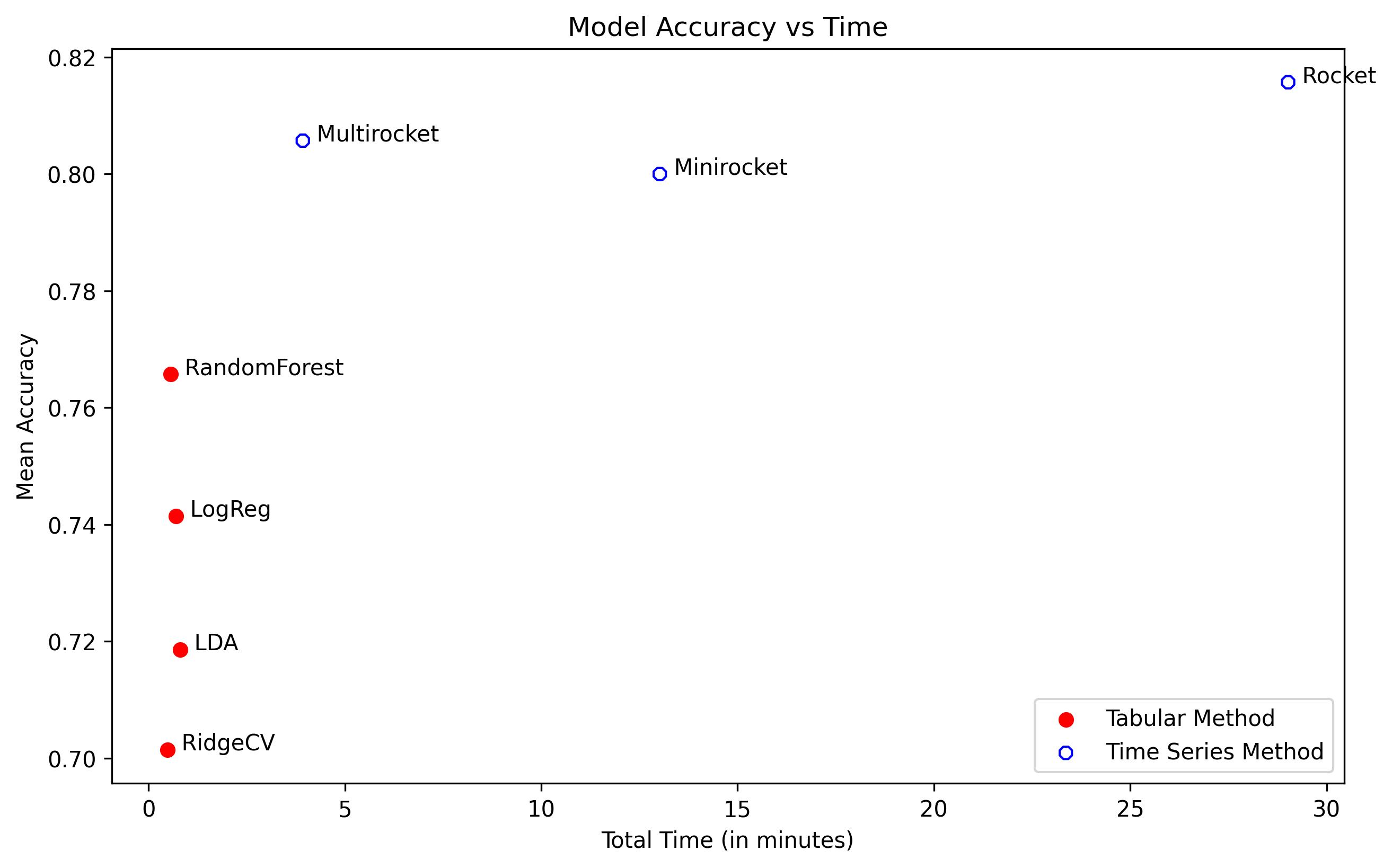}
    \caption{Accuracy-Time tradeoff for datasets in the \textbf{grey region} in Figure \ref{fig:lm_ts_mtsc}.}
    \label{fig:acc_time_tradeoff_mtsc}
\end{figure}
\subsubsection{Computation Time Analysis}

For the same reasons as for the univariate time series classification task, we perform the time-accuracy tradeoff analysis for multivariate time series classification. Figure \ref{fig:acc_time_tradeoff_mtsc} illustrates the performance of various time-series and tabular models on the datasets in the grey region of Figure \ref{fig:lm_ts_mtsc}. Rocket is the most accurate among time-series models, and Random Forest is the most accurate model among tabular models. The difference between the mean accuracy of Rocket and the mean accuracy of Random Forest is about 5 percentage point, while the difference in total computation time is about 4 minutes.

In addition to considering the trade-off between time and accuracy, we also analyzed the domain-wise performance of tabular and time series models in multivariate datasets in Table \ref{tab:mtsc_domain}. The datasets consisted of 6 domains, with 60\% of the data coming from two domains (HAR and EEG). Time series models generally performed well, but tabular models performed better in the ECG and EEG/MEG domains.

\begin{table}[H]
    \centering
\caption{Mean accuracy of classifiers by problem types on UCR multivariate datasets.}
\begin{tabular}{c|cccc|ccc}
\hline
& \multicolumn{4}{c|}{Tabular Models} & \multicolumn{3}{c}{Time Series Models}\\
\hline
Domain (\#datasets)   &   RidgeCV &   LDA &   LogReg &   RandomForest &   Rocket &   Minirocket &   Multirocket \\
\hline
 HAR(9) &      0.67 &   0.53 &      0.74 &          \textbf{0.78} &     0.92 &         \textbf{0.94} &          0.94 \\
 EEG/MEG(6) &      0.55 &   0.54 &      \textbf{0.58} &          0.50  &     \textbf{0.55} &         0.55 &          0.54   \\
 Audio Spectra(3) &      \textbf{0.18} &   0.16 &      0.18 &          0.18 &     0.46 &         \textbf{0.70}  &          0.52  \\
 Other(3) &      0.52 &   0.58 &      0.65 &          \textbf{0.74} &     \textbf{0.84} &         0.83 &          0.80   \\
 ECG(2) &      \textbf{0.46} &   0.20  &      \textbf{0.46} &          0.40  &     \textbf{0.27} &         0.26 &          0.24  \\
 Motion(2) &      0.59 &   0.65 &      0.65 &          \textbf{0.72} &     0.99 &         \textbf{1.00}    &          \textbf{1.00}     \\
\hline
\end{tabular}
    \label{tab:mtsc_domain}
\end{table}




\subsection{Discussion and Lessons Learned}
\begin{itemize}
    \item \textbf{Redefining baselines:} Most previous research has considered 1NN-DTW as the baseline for time series classification. This is a reasonable choice, as 1NN-DTW is a simple and effective algorithm that is often competitive with more complex time series methods. However, our study suggests that simple tabular models can perform significantly well on some datasets, even when compared to recent state-of-the-art TSC algorithms. This finding suggests that there is a  need to rethink how we do baseline comparisons for time series classification.
    
    \item \textbf{Not all that looks time series is a time series:}
    Our study demonstrated that tabular methods outperformed time series methods on some domains, specifically Spectro (Table \ref{tab:utsc_domain}), EEG or ECG (Table \ref{tab:mtsc_domain}). This could be because the Spectro datasets did not contain strong temporal information. Either way, we need to ask whether it makes sense to have these datasets in a time series classification benchmark.

    \item \textbf{Considering trade-offs:} In our study we observed that time series models outperformed tabular models by a few percentage points on the red datasets. However, tabular models outperformed time series methods in the green datasets and were significantly faster to train and test. Therefore, especially for datasets in the grey region, where tabular and time series methods are close in accuracy, we recommend carefully considering whether tabular models are preferable to time series methods, especially if time is a constraint.
\end{itemize}

\subsection{Improving Tabular Models}
Since the above-mentioned experiments were conducted using the default hyperparameters, we wanted to investigate whether we could improve the performance of tabular models by tuning the hyperparameters. To do this, we performed hyperparameter tuning on Random Forest and Logistic Regression, since they were the best performing models in both univariate (Figure \ref{fig:acc_time_tradeoff_uts}) and multivariate (Figure \ref{fig:acc_time_tradeoff_mtsc}) experiments.

We performed hyperparameter tuning with a combination of scaling and regularization. Table \ref{tab:hyper} shows the results of the hyperparameter tuning and the improvement for the best tabular model. We found that hyperparameter tuning can increase accuracy, but it also takes a significant amount of time to find the best hyperparameters.


\begin{table}[]
    \centering
    \caption{Improvement on accuracy on univariate and multivariate datasets and mean computation time in minutes.}
    \begin{tabular}{c|cc|cc}
    \hline
         &  \multicolumn{2}{c|}{Mean Accuracy}  &  \multicolumn{2}{c}{Mean Computation Time (minutes)} \\
         \hline
         & Before & After & Before & After \\
        \hline
    Univariate  & 0.86  & 0.87 & 0.47 & 13.41 \\ 
    Multivariate  & 0.74 & 0.75 & 0.91 & 43.10 \\
    \hline
    \end{tabular}
    \label{tab:hyper}
\end{table}

\section{Conclusion}

In this study, we compared the performance of tabular models with state-of-the-art time series models on the UCR/UEA univariate and multivariate time series classification benchmarks. We found that tabular models performed surprisingly well on many datasets, outperforming the recent Multirocket classifier on a significant percentage of the datasets. On many other datasets, the accuracy was comparable, but tabular models were more efficient in terms of computation time. Overall, in about half of the datasets in either the univariate or the multivariate benchmarks, tabular methods were within 10 percentage points accuracy of the time series methods.

Our findings suggest that tabular models should be considered as baselines for evaluating improvements in time series classifiers, and even for considering whether a dataset should be included in the time series classification benchmarks. Furthermore, tabular methods can be a viable alternative to time series models for some classification tasks. Tabular models are easier to train and deploy, and they are more efficient in terms of computation time. The performance of tabular models does vary depending on the characteristics of the dataset. In future work, we plan to further investigate the factors that contribute to the performance of tabular models on time series data, and include more tabular  models and parameter tuning.
\section*{Acknowledgement}
This publication has emanated from research supported in part by a grant from Science Foundation Ireland through the VistaMilk SFI Research Centre \\ 
(SFI/16/RC/3835) and the Insight Centre for Data Analytics (12/RC/2289 P2). For the purpose of Open Access, the author has applied a CC BY public copyright licence to any Author Accepted Manuscript version arising from this submission. We would like to thank the reviewers for their constructive feedback. We would also like to thank all the researchers that have contributed open source code and datasets to the UEA MTSC Archive and especially, we want to thank the groups at UEA and UCR who continue to maintain and expand the archive.

\bibliographystyle{unsrt}
\bibliography{ref}
\newpage
\section*{Appendix}

\begin{table}[!h]
    \centering
    \small
        \caption{Data dictionary for Multivariate time series classification.}
\begin{tabular}{l|ccccccc}
\hline
Domain & Datasets  & \rot{Train Size} & \rot{Test Size} & \rot{\#Channels} & \rot{TS-len} & \rot{\#Classes} \\
\hline
Audio Spectra  & DuckDuckGeese &  50 & 50 & 1345 & 270 & 5 \\
Other & PEMS-SF & 267 & 173 & 963 & 144 & 7 \\
EEG/MEG  & FaceDetection  & 5890 & 3524 & 144 & 62 & 2 \\
EEG/MEG  & MotorImagery  & 278 & 100 & 64 & 3000 & 2 \\
Audio Spectra  & Heartbeat  & 204 & 205 & 61 & 405 & 2 \\
EEG/MEG  & FingerMovements  & 316 & 100 & 28 & 50 & 2 \\
Human Activity Recogntion & NATOPS & 180 & 180 & 24 & 51 & 6 \\
Audio Spectra  & PhonemeSpectra  & 3315 & 3353 & 11 & 217 & 39 \\
EEG/MEG  & HandMovementDirection  & 160 & 74 & 10 & 400 & 4 \\
Motion  & ArticularyWordRecognition  & 275 & 300 & 9 & 144 & 25 \\
EEG/MEG  & SelfRegulationSCP2  & 200 & 180 & 7 & 1152 & 2 \\
EEG/MEG  & SelfRegulationSCP1  & 268 & 293 & 6 & 896 & 2 \\
Human Activity Recogntion & BasicMotions  & 40 & 40 & 6 & 100 & 4 \\
Human Activity Recogntion & Cricket  & 108 & 72 & 6 & 1197 & 12 \\
Human Activity Recogntion & EigenWorms  & 128 & 131 & 6 & 17984 & 5 \\
Human Activity Recogntion & LSST &  2459 & 2466 & 6 & 36 & 14 \\
Human Activity Recogntion & RacketSports  & 151 & 152 & 6 & 30 & 4 \\
ECG  & StandWalkJump &  12 & 15 & 4 & 2500 & 3 \\
Human Activity Recogntion & ERing &  30 & 270 & 4 & 65 & 6 \\
Human Activity Recogntion & Handwriting  & 150 & 850 & 3 & 152 & 26 \\
Human Activity Recogntion & UWaveGestureLibrary  & 120 & 320 & 3 & 315 & 8 \\
Motion  & Epilepsy &  137 & 138 & 3 & 206 & 4 \\
Other & EthanolConcentration &  261 & 263 & 3 & 1751 & 4 \\
ECG  & AtrialFibrillation  & 15 & 15 & 2 & 640 & 3 \\
Motion  & PenDigits &  7494 & 3498 & 2 & 8 & 10 \\
Other & Libras & 180 & 180 & 2 & 45 & 15 \\
\hline
\end{tabular}
    \label{tab:ddic_mtsc}
\end{table}

\renewcommand{\arraystretch}{1}
 \begin{longtable}[!h]{lc|ccccc}
\caption{Data dictionary for Univariate time series classification.} \\
\hline
Data & \rot{Train Size} & \rot{Test Size} & \rot{TS-Len} & \rot{\#Classes} & \rot{Domain} \\
\hline
ACSF1 & 100 & 100 & 1460 & 10 & DEVICE \\
Adiac & 390 & 391 & 176 & 37 & IMAGE \\
ArrowHead & 36 & 175 & 251 & 3 & IMAGE \\
Beef & 30 & 30 & 470 & 5 & SPECTRO \\
BeetleFly & 20 & 20 & 512 & 2 & IMAGE \\
BirdChicken & 20 & 20 & 512 & 2 & IMAGE \\
BME & 30 & 150 & 128 & 3 & SIMULATED \\
Car & 60 & 60 & 577 & 4 & SENSOR \\
CBF & 30 & 900 & 128 & 3 & SIMULATED \\
Chinatown & 20 & 345 & 24 & 2 & Traffic \\
ChlorineConcentration & 467 & 3840 & 166 & 3 & SIMULATED \\
CinCECGTorso & 40 & 1380 & 1639 & 4 & ECG \\
Coffee & 28 & 28 & 286 & 2 & SPECTRO \\
Computers & 250 & 250 & 720 & 2 & DEVICE \\
CricketX & 390 & 390 & 300 & 12 & MOTION \\
CricketY & 390 & 390 & 300 & 12 & MOTION \\
CricketZ & 390 & 390 & 300 & 12 & MOTION \\
Crop & 7200 & 16800 & 46 & 24 & IMAGE \\
DiatomSizeReduction & 16 & 306 & 345 & 4 & IMAGE \\
DistalPhalanxOutlineAgeGroup & 400 & 139 & 80 & 3 & IMAGE \\
DistalPhalanxOutlineCorrect & 600 & 276 & 80 & 2 & IMAGE \\
DistalPhalanxTW & 400 & 139 & 80 & 6 & IMAGE \\
Earthquakes & 322 & 139 & 512 & 2 & SENSOR \\
ECG200 & 100 & 100 & 96 & 2 & ECG \\
ECG5000 & 500 & 4500 & 140 & 5 & ECG \\
ECGFiveDays & 23 & 861 & 136 & 2 & ECG \\
ElectricDevices & 8926 & 7711 & 96 & 7 & DEVICE \\
EOGHorizontalSignal & 362 & 362 & 1250 & 12 & EOG \\
EOGVerticalSignal & 362 & 362 & 1250 & 12 & EOG \\
EthanolLevel & 504 & 500 & 1751 & 4 & SPECTRO \\
FaceAll & 560 & 1690 & 131 & 14 & IMAGE \\
FaceFour & 24 & 88 & 350 & 4 & IMAGE \\
FacesUCR & 200 & 2050 & 131 & 14 & IMAGE \\
FiftyWords & 450 & 455 & 270 & 50 & IMAGE \\
Fish & 175 & 175 & 463 & 7 & IMAGE \\
FordA & 3601 & 1320 & 500 & 2 & SENSOR \\
FordB & 3636 & 810 & 500 & 2 & SENSOR \\
FreezerRegularTrain & 150 & 2850 & 301 & 2 & SENSOR \\
FreezerSmallTrain & 28 & 2850 & 301 & 2 & SENSOR \\
GunPoint & 50 & 150 & 150 & 2 & MOTION \\
GunPointAgeSpan & 135 & 316 & 150 & 2 & MOTION \\
GunPointMaleVersusFemale & 135 & 316 & 150 & 2 & MOTION \\
GunPointOldVersusYoung & 135 & 316 & 150 & 2 & MOTION \\
Ham & 109 & 105 & 431 & 2 & SPECTRO \\
Haptics & 155 & 308 & 1092 & 5 & MOTION \\
Herring & 64 & 64 & 512 & 2 & IMAGE \\
HouseTwenty & 34 & 101 & 3000 & 2 & DEVICE \\
InlineSkate & 100 & 550 & 1882 & 7 & MOTION \\
InsectEPGRegularTrain & 62 & 249 & 601 & 3 & EPG \\
InsectEPGSmallTrain & 17 & 249 & 601 & 3 & EPG \\
ItalyPowerDemand & 67 & 1029 & 24 & 2 & SENSOR \\
LargeKitchenAppliances & 375 & 375 & 720 & 3 & DEVICE \\
Lightning2 & 60 & 61 & 637 & 2 & SENSOR \\
Lightning7 & 70 & 73 & 319 & 7 & SENSOR \\
Mallat & 55 & 2345 & 1024 & 8 & SIMULATED \\
Meat & 60 & 60 & 448 & 3 & SPECTRO \\
MedicalImages & 381 & 760 & 99 & 10 & IMAGE \\
MiddlePhalanxOutlineAgeGroup & 400 & 154 & 80 & 3 & IMAGE \\
MiddlePhalanxOutlineCorrect & 600 & 291 & 80 & 2 & IMAGE \\
MiddlePhalanxTW & 399 & 154 & 80 & 6 & IMAGE \\
MixedShapes & 500 & 2425 & 1024 & 5 & IMAGE \\
MixedShapesSmallTrain & 100 & 2425 & 1024 & 5 & IMAGE \\
MoteStrain & 20 & 1252 & 84 & 2 & SENSOR \\
OliveOil & 30 & 30 & 570 & 4 & SPECTRO \\
OSULeaf & 200 & 242 & 427 & 6 & IMAGE \\
PhalangesOutlinesCorrect & 1800 & 858 & 80 & 2 & IMAGE \\
Phoneme & 214 & 1896 & 1024 & 39 & SOUND \\
PigAirwayPressure & 104 & 208 & 2000 & 52 & HEMODYNAMICS \\
PigArtPressure & 104 & 208 & 2000 & 52 & HEMODYNAMICS \\
PigCVP & 104 & 208 & 2000 & 52 & HEMODYNAMICS \\
Plane & 105 & 105 & 144 & 7 & SENSOR \\
PowerCons & 180 & 180 & 144 & 2 & DEVICE \\
ProximalPhalanxOutlineAgeGroup & 400 & 205 & 80 & 3 & IMAGE \\
ProximalPhalanxOutlineCorrect & 600 & 291 & 80 & 2 & IMAGE \\
ProximalPhalanxTW & 400 & 205 & 80 & 6 & IMAGE \\
RefrigerationDevices & 375 & 375 & 720 & 3 & DEVICE \\
Rock & 20 & 50 & 2844 & 4 & SPECTRO \\
ScreenType & 375 & 375 & 720 & 3 & DEVICE \\
SemgHandGenderCh2 & 300 & 600 & 1500 & 2 & SPECTRO \\
SemgHandMovementCh2 & 450 & 450 & 1500 & 6 & SPECTRO \\
SemgHandSubjectCh2 & 450 & 450 & 1500 & 5 & SPECTRO \\
ShapeletSim & 20 & 180 & 500 & 2 & SIMULATED \\
ShapesAll & 600 & 600 & 512 & 60 & IMAGE \\
SmallKitchenAppliances & 375 & 375 & 720 & 3 & DEVICE \\
SmoothSubspace & 150 & 150 & 15 & 3 & SIMULATED \\
SonyAIBORobotSurface1 & 20 & 601 & 70 & 2 & SENSOR \\
SonyAIBORobotSurface2 & 27 & 953 & 65 & 2 & SENSOR \\
StarLightCurves & 1000 & 8236 & 1024 & 3 & SENSOR \\
Strawberry & 613 & 370 & 235 & 2 & SPECTRO \\
SwedishLeaf & 500 & 625 & 128 & 15 & IMAGE \\
Symbols & 25 & 995 & 398 & 6 & IMAGE \\
SyntheticControl & 300 & 300 & 60 & 6 & SIMULATED \\
ToeSegmentation1 & 40 & 228 & 277 & 2 & MOTION \\
ToeSegmentation2 & 36 & 130 & 343 & 2 & MOTION \\
Trace & 100 & 100 & 275 & 4 & SENSOR \\
TwoLeadECG & 23 & 1139 & 82 & 2 & ECG \\
TwoPatterns & 1000 & 4000 & 128 & 4 & SIMULATED \\
UMD & 36 & 144 & 150 & 3 & SIMULATED \\
UWaveGestureLibraryAll & 896 & 3582 & 945 & 8 & MOTION \\
UWaveGestureLibraryX & 896 & 3582 & 315 & 8 & MOTION \\
UWaveGestureLibraryY & 896 & 3582 & 315 & 8 & MOTION \\
UWaveGestureLibraryZ & 896 & 3582 & 315 & 8 & MOTION \\
Wafer & 1000 & 6164 & 152 & 2 & SENSOR \\
Wine & 57 & 54 & 234 & 2 & SPECTRO \\
\hline
\label{tab:ddic_univariate}
\end{longtable}

\renewcommand{\arraystretch}{1}
\begin{longtable}{l|cccc|ccc}
\caption{Accuracy of tabular and time series methods on UTSC datasets.} \\
\hline
Name & \rot{RidgeCV} & \rot{LDA} & \rot{LogRegCV} & \rot{RandomForest}  & \rot{Rocket} & \rot{Minirocket} & \rot{Multirocket} \\
\hline
ACSF1 & 0.42 & 0.41 & 0.62 & 0.75 & 0.90 & 0.91 & 0.88 \\
Adiac & 0.44 & 0.53 & 0.73 & 0.65 & 0.79 & 0.83 & 0.83 \\
ArrowHead & 0.73 & 0.67 & 0.73 & 0.70 & 0.82 & 0.84 & 0.87 \\
Beef & 0.87 & 0.93 & 0.87 & 0.77 & 0.83 & 0.87 & 0.77 \\
BeetleFly & 0.85 & 0.75 & 0.85 & 0.85 & 0.90 & 0.90 & 0.85 \\
BirdChicken & 0.50 & 0.55 & 0.70 & 0.75 & 0.90 & 0.90 & 0.90 \\
BME & 0.91 & 0.95 & 0.91 & 0.97 & 1.00 & 1.00 & 1.00 \\
Car & 0.80 & 0.80 & 0.83 & 0.67 & 0.90 & 0.92 & 0.92 \\
CBF & 0.83 & 0.84 & 0.85 & 0.89 & 1.00 & 1.00 & 1.00 \\
Chinatown & 0.98 & 0.95 & 0.98 & 0.98 & 0.98 & 0.98 & 0.98 \\
ChlorineConcentration & 0.85 & 0.88 & 0.78 & 0.71 & 0.82 & 0.77 & 0.79 \\
CinCECGTorso & 0.39 & 0.45 & 0.45 & 0.72 & 0.83 & 0.87 & 0.95 \\
Coffee & 1.00 & 1.00 & 1.00 & 0.96 & 1.00 & 1.00 & 1.00 \\
Computers & 0.51 & 0.49 & 0.48 & 0.62 & 0.75 & 0.70 & 0.78 \\
CricketX & 0.27 & 0.13 & 0.27 & 0.55 & 0.82 & 0.81 & 0.81 \\
CricketY & 0.37 & 0.15 & 0.39 & 0.60 & 0.86 & 0.83 & 0.85 \\
CricketZ & 0.31 & 0.15 & 0.28 & 0.57 & 0.85 & 0.82 & 0.84 \\
Crop & 0.56 & 0.63 & 0.69 & 0.76 & 0.75 & 0.75 & 0.77 \\
DiatomSizeReduction & 0.96 & 0.96 & 0.95 & 0.90 & 0.98 & 0.92 & 0.96 \\
DistalPhalanxOutlineAgeGroup & 0.66 & 0.60 & 0.69 & 0.77 & 0.76 & 0.75 & 0.78 \\
DistalPhalanxOutlineCorrect & 0.66 & 0.66 & 0.65 & 0.76 & 0.76 & 0.79 & 0.79 \\
DistalPhalanxTW & 0.61 & 0.58 & 0.60 & 0.68 & 0.72 & 0.70 & 0.69 \\
Earthquakes & 0.75 & 0.65 & 0.68 & 0.75 & 0.75 & 0.75 & 0.75 \\
ECG200 & 0.80 & 0.59 & 0.84 & 0.83 & 0.91 & 0.91 & 0.92 \\
ECG5000 & 0.93 & 0.93 & 0.94 & 0.94 & 0.95 & 0.95 & 0.95 \\
ECGFiveDays & 0.99 & 0.94 & 0.97 & 0.80 & 1.00 & 1.00 & 1.00 \\
ElectricDevices & 0.44 & 0.46 & 0.47 & 0.65 & 0.73 & 0.73 & 0.73 \\
EOGHorizontalSignal & 0.34 & 0.27 & 0.39 & 0.44 & 0.64 & 0.59 & 0.65 \\
EOGVerticalSignal & 0.25 & 0.28 & 0.35 & 0.42 & 0.54 & 0.56 & 0.54 \\
EthanolLevel & 0.66 & 0.91 & 0.72 & 0.48 & 0.57 & 0.61 & 0.62 \\
FaceAll & 0.79 & 0.79 & 0.77 & 0.79 & 0.95 & 0.81 & 0.80 \\
FaceFour & 0.89 & 0.85 & 0.86 & 0.75 & 0.97 & 0.99 & 0.94 \\
FacesUCR & 0.70 & 0.62 & 0.73 & 0.77 & 0.96 & 0.96 & 0.96 \\
FiftyWords & 0.43 & 0.32 & 0.56 & 0.63 & 0.83 & 0.84 & 0.86 \\
Fish & 0.82 & 0.73 & 0.85 & 0.77 & 0.98 & 0.99 & 0.98 \\
FordA & 0.52 & 0.53 & 0.49 & 0.74 & 0.94 & 0.95 & 0.95 \\
FordB & 0.50 & 0.50 & 0.49 & 0.63 & 0.79 & 0.81 & 0.83 \\
FreezerRegularTrain & 0.99 & 0.98 & 0.98 & 0.95 & 1.00 & 1.00 & 1.00 \\
FreezerSmallTrain & 0.86 & 0.94 & 0.81 & 0.75 & 0.95 & 0.97 & 0.99 \\
GunPoint & 0.85 & 0.81 & 0.85 & 0.92 & 1.00 & 0.99 & 1.00 \\
GunPointAgeSpan & 0.87 & 0.57 & 0.89 & 0.97 & 1.00 & 0.99 & 1.00 \\
GunPointMaleVersusFemale & 0.97 & 0.68 & 0.99 & 0.97 & 1.00 & 1.00 & 1.00 \\
GunPointOldVersusYoung & 1.00 & 0.88 & 1.00 & 1.00 & 0.99 & 1.00 & 1.00 \\
Ham & 0.71 & 0.66 & 0.65 & 0.75 & 0.71 & 0.69 & 0.73 \\
Haptics & 0.43 & 0.35 & 0.38 & 0.44 & 0.52 & 0.53 & 0.56 \\
Herring & 0.59 & 0.58 & 0.63 & 0.66 & 0.70 & 0.66 & 0.67 \\
HouseTwenty & 0.73 & 0.72 & 0.72 & 0.71 & 0.97 & 0.97 & 0.99 \\
InlineSkate & 0.19 & 0.23 & 0.27 & 0.34 & 0.46 & 0.45 & 0.47 \\
InsectEPGRegularTrain & 0.82 & 1.00 & 1.00 & 1.00 & 1.00 & 1.00 & 1.00 \\
InsectEPGSmallTrain & 0.83 & 1.00 & 1.00 & 1.00 & 0.98 & 1.00 & 1.00 \\
InsectWingbeatSound & 0.62 & 0.26 & 0.58 & 0.63 & 0.66 & 0.67 & 0.68 \\
ItalyPowerDemand & 0.97 & 0.94 & 0.96 & 0.97 & 0.97 & 0.96 & 0.97 \\
LargeKitchenAppliances & 0.44 & 0.38 & 0.39 & 0.58 & 0.90 & 0.87 & 0.88 \\
Lightning2 & 0.77 & 0.66 & 0.72 & 0.75 & 0.75 & 0.74 & 0.69 \\
Lightning7 & 0.64 & 0.55 & 0.67 & 0.71 & 0.84 & 0.79 & 0.82 \\
Mallat & 0.76 & 0.86 & 0.82 & 0.91 & 0.96 & 0.95 & 0.92 \\
Meat & 0.98 & 0.98 & 0.93 & 0.92 & 0.95 & 0.97 & 0.93 \\
MedicalImages & 0.55 & 0.49 & 0.63 & 0.73 & 0.80 & 0.80 & 0.81 \\
MiddlePhalanxOutlineAgeGroup & 0.60 & 0.48 & 0.60 & 0.62 & 0.60 & 0.60 & 0.62 \\
MiddlePhalanxOutlineCorrect & 0.62 & 0.58 & 0.59 & 0.81 & 0.83 & 0.84 & 0.86 \\
MiddlePhalanxTW & 0.61 & 0.53 & 0.53 & 0.56 & 0.55 & 0.53 & 0.54 \\
MixedShapes & 0.79 & 0.71 & 0.82 & 0.87 & 0.97 & 0.97 & 0.98 \\
MixedShapesSmallTrain & 0.77 & 0.69 & 0.78 & 0.78 & 0.94 & 0.95 & 0.96 \\
MoteStrain & 0.86 & 0.72 & 0.86 & 0.88 & 0.91 & 0.93 & 0.95 \\
OliveOil & 0.90 & 0.90 & 0.90 & 0.90 & 0.90 & 0.93 & 0.97 \\
OSULeaf & 0.40 & 0.32 & 0.46 & 0.49 & 0.93 & 0.96 & 0.96 \\
PhalangesOutlinesCorrect & 0.67 & 0.66 & 0.67 & 0.82 & 0.83 & 0.84 & 0.85 \\
Phoneme & 0.11 & 0.08 & 0.10 & 0.13 & 0.28 & 0.27 & 0.35 \\
PigAirwayPressure & 0.02 & 0.21 & 0.08 & 0.09 & 0.09 & 0.88 & 0.60 \\
PigArtPressure & 0.10 & 0.12 & 0.17 & 0.19 & 0.95 & 0.99 & 0.95 \\
PigCVP & 0.04 & 0.14 & 0.10 & 0.11 & 0.93 & 0.95 & 0.88 \\
Plane & 0.98 & 0.99 & 0.98 & 0.98 & 1.00 & 1.00 & 1.00 \\
PowerCons & 0.98 & 0.73 & 0.99 & 1.00 & 0.92 & 0.99 & 0.98 \\
ProximalPhalanxOutlineAgeGroup & 0.84 & 0.83 & 0.85 & 0.86 & 0.85 & 0.85 & 0.86 \\
ProximalPhalanxOutlineCorrect & 0.84 & 0.84 & 0.85 & 0.86 & 0.90 & 0.91 & 0.91 \\
ProximalPhalanxTW & 0.75 & 0.75 & 0.76 & 0.80 & 0.81 & 0.82 & 0.82 \\
RefrigerationDevices & 0.35 & 0.35 & 0.37 & 0.53 & 0.53 & 0.48 & 0.50 \\
Rock & 0.88 & 0.94 & 0.84 & 0.66 & 0.90 & 0.80 & 0.86 \\
ScreenType & 0.44 & 0.40 & 0.39 & 0.42 & 0.49 & 0.47 & 0.57 \\
SemgHandGenderCh2 & 0.85 & 0.76 & 0.82 & 0.85 & 0.92 & 0.90 & 0.96 \\
SemgHandMovementCh2 & 0.50 & 0.39 & 0.50 & 0.50 & 0.62 & 0.71 & 0.78 \\
SemgHandSubjectCh2 & 0.78 & 0.59 & 0.81 & 0.68 & 0.89 & 0.87 & 0.92 \\
ShapeletSim & 0.49 & 0.51 & 0.48 & 0.51 & 1.00 & 1.00 & 1.00 \\
ShapesAll & 0.50 & 0.11 & 0.63 & 0.73 & 0.91 & 0.92 & 0.93 \\
SmallKitchenAppliances & 0.54 & 0.35 & 0.41 & 0.71 & 0.81 & 0.82 & 0.82 \\
SmoothSubspace & 0.80 & 0.83 & 0.86 & 0.99 & 0.98 & 0.94 & 0.98 \\
SonyAIBORobotSurface1 & 0.69 & 0.70 & 0.68 & 0.67 & 0.92 & 0.89 & 0.89 \\
SonyAIBORobotSurface2 & 0.83 & 0.81 & 0.81 & 0.81 & 0.91 & 0.92 & 0.94 \\
StarLightCurves & 0.85 & 0.81 & 0.92 & 0.95 & 0.98 & 0.98 & 0.98 \\
Strawberry & 0.93 & 0.95 & 0.95 & 0.96 & 0.98 & 0.98 & 0.98 \\
SwedishLeaf & 0.66 & 0.72 & 0.83 & 0.87 & 0.97 & 0.97 & 0.98 \\
Symbols & 0.77 & 0.82 & 0.82 & 0.85 & 0.97 & 0.98 & 0.98 \\
SyntheticControl & 0.80 & 0.93 & 0.91 & 0.96 & 1.00 & 0.98 & 1.00 \\
ToeSegmentation1 & 0.57 & 0.55 & 0.58 & 0.62 & 0.96 & 0.96 & 0.95 \\
ToeSegmentation2 & 0.55 & 0.54 & 0.56 & 0.75 & 0.92 & 0.92 & 0.92 \\
Trace & 0.61 & 0.70 & 0.76 & 0.83 & 1.00 & 1.00 & 1.00 \\
TwoLeadECG & 0.94 & 0.89 & 0.95 & 0.73 & 1.00 & 1.00 & 1.00 \\
TwoPatterns & 0.79 & 0.84 & 0.84 & 0.83 & 1.00 & 1.00 & 1.00 \\
UMD & 0.82 & 0.79 & 0.84 & 0.95 & 0.99 & 0.99 & 0.99 \\
UWaveGestureLibraryAll & 0.85 & 0.28 & 0.81 & 0.93 & 0.98 & 0.97 & 0.98 \\
UWaveGestureLibraryX & 0.63 & 0.51 & 0.63 & 0.76 & 0.86 & 0.85 & 0.87 \\
UWaveGestureLibraryY & 0.53 & 0.42 & 0.58 & 0.68 & 0.77 & 0.78 & 0.80 \\
UWaveGestureLibraryZ & 0.51 & 0.45 & 0.55 & 0.71 & 0.79 & 0.80 & 0.82 \\
Wafer & 0.94 & 0.94 & 0.94 & 0.99 & 1.00 & 1.00 & 1.00 \\
Wine & 0.83 & 0.91 & 0.89 & 0.78 & 0.81 & 0.83 & 0.89 \\
WordSynonyms & 0.38 & 0.23 & 0.46 & 0.55 & 0.75 & 0.76 & 0.78 \\
Worms & 0.38 & 0.42 & 0.34 & 0.55 & 0.74 & 0.75 & 0.75 \\
WormsTwoClass & 0.55 & 0.62 & 0.52 & 0.62 & 0.81 & 0.77 & 0.78 \\
Yoga & 0.65 & 0.59 & 0.67 & 0.81 & 0.91 & 0.91 & 0.92 \\
\hline
\label{tab:acc_utsc}
\end{longtable}

\begin{longtable}{l|cccc|ccc}
\caption{Computation time (in minutes) for univariate datasets.} \\
\hline
Name & \rot{RidgeCV} & \rot{LDA} & \rot{LogReg} & \rot{RandomForest} & \rot{Rocket} & \rot{Minirocket} & \rot{Multirocket} \\
\hline
ACSF1 & 0.03 & 0.04 & 0.29 & 0.19 & 0.83 & 0.12 & 0.26 \\
Adiac & 0.03 & 0.02 & 0.12 & 0.40 & 0.38 & 0.07 & 0.17 \\
ArrowHead & 0.01 & 0.01 & 0.03 & 0.11 & 0.14 & 0.02 & 0.07 \\
Beef & 0.01 & 0.01 & 0.05 & 0.10 & 0.08 & 0.02 & 0.04 \\
BeetleFly & 0.01 & 0.01 & 0.02 & 0.09 & 0.06 & 0.01 & 0.04 \\
BirdChicken & 0.01 & 0.01 & 0.04 & 0.09 & 0.06 & 0.02 & 0.04 \\
BME & 0.01 & 0.00 & 0.01 & 0.09 & 0.07 & 0.01 & 0.04 \\
Car & 0.01 & 0.01 & 0.06 & 0.11 & 0.19 & 0.04 & 0.09 \\
CBF & 0.01 & 0.00 & 0.01 & 0.09 & 0.31 & 0.05 & 0.18 \\
Chinatown & 0.01 & 0.00 & 0.01 & 0.09 & 0.03 & 0.01 & 0.03 \\
ChlorineConcentration & 0.02 & 0.02 & 0.04 & 0.40 & 1.85 & 0.28 & 1.08 \\
CinCECGTorso & 0.03 & 0.03 & 0.15 & 0.14 & 5.69 & 0.90 & 2.35 \\
Coffee & 0.01 & 0.01 & 0.02 & 0.09 & 0.04 & 0.02 & 0.03 \\
Computers & 0.03 & 0.05 & 0.04 & 0.28 & 0.87 & 0.21 & 0.37 \\
CricketX & 0.04 & 0.03 & 0.11 & 0.36 & 0.57 & 0.13 & 0.31 \\
CricketY & 0.03 & 0.03 & 0.11 & 0.34 & 0.57 & 0.15 & 0.31 \\
CricketZ & 0.03 & 0.03 & 0.13 & 0.37 & 0.57 & 0.13 & 0.31 \\
Crop & 0.08 & 0.05 & 0.53 & 2.98 & 4.23 & 1.86 & 5.57 \\
DiatomSizeReduction & 0.01 & 0.01 & 0.04 & 0.09 & 0.29 & 0.05 & 0.12 \\
DistalPhalanxOutlineAgeGroup & 0.01 & 0.01 & 0.03 & 0.18 & 0.13 & 0.04 & 0.09 \\
DistalPhalanxOutlineCorrect & 0.01 & 0.01 & 0.02 & 0.26 & 0.20 & 0.05 & 0.13 \\
DistalPhalanxTW & 0.01 & 0.01 & 0.03 & 0.18 & 0.12 & 0.04 & 0.08 \\
Earthquakes & 0.03 & 0.04 & 0.02 & 0.29 & 0.58 & 0.13 & 0.35 \\
ECG200 & 0.01 & 0.01 & 0.01 & 0.11 & 0.05 & 0.02 & 0.04 \\
ECG5000 & 0.02 & 0.02 & 0.05 & 0.26 & 1.66 & 0.32 & 0.83 \\
ECGFiveDays & 0.01 & 0.00 & 0.01 & 0.09 & 0.28 & 0.05 & 0.14 \\
ElectricDevices & 0.24 & 0.10 & 0.15 & 6.60 & 6.03 & 2.79 & 6.49 \\
EOGHorizontalSignal & 0.09 & 0.14 & 0.47 & 0.51 & 2.19 & 0.44 & 0.94 \\
EOGVerticalSignal & 0.06 & 0.14 & 0.44 & 0.53 & 2.18 & 0.42 & 0.95 \\
EthanolLevel & 0.16 & 0.30 & 0.41 & 0.86 & 4.29 & 0.81 & 1.66 \\
FaceAll & 0.02 & 0.02 & 0.13 & 0.40 & 0.69 & 0.13 & 0.36 \\
FaceFour & 0.02 & 0.01 & 0.04 & 0.11 & 0.10 & 0.02 & 0.06 \\
FacesUCR & 0.02 & 0.01 & 0.06 & 0.20 & 0.67 & 0.14 & 0.34 \\
FiftyWords & 0.05 & 0.04 & 0.46 & 0.73 & 0.58 & 0.13 & 0.24 \\
Fish & 0.02 & 0.03 & 0.15 & 0.21 & 0.38 & 0.08 & 0.16 \\
FordA & 0.58 & 0.45 & 0.19 & 5.25 & 6.03 & 1.44 & 2.77 \\
FordB & 0.78 & 0.34 & 0.23 & 5.80 & 5.51 & 1.34 & 2.62 \\
FreezerRegularTrain & 0.02 & 0.03 & 0.02 & 0.16 & 2.03 & 0.34 & 0.98 \\
FreezerSmallTrain & 0.01 & 0.01 & 0.03 & 0.12 & 1.93 & 0.31 & 0.92 \\
GunPoint & 0.01 & 0.01 & 0.03 & 0.11 & 0.08 & 0.02 & 0.04 \\
GunPointAgeSpan & 0.01 & 0.01 & 0.02 & 0.13 & 0.16 & 0.03 & 0.08 \\
GunPointMaleVersusFemale & 0.01 & 0.01 & 0.01 & 0.12 & 0.16 & 0.03 & 0.08 \\
GunPointOldVersusYoung & 0.01 & 0.01 & 0.01 & 0.11 & 0.16 & 0.03 & 0.08 \\
Ham & 0.02 & 0.03 & 0.02 & 0.14 & 0.22 & 0.05 & 0.10 \\
Haptics & 0.03 & 0.05 & 0.14 & 0.24 & 1.17 & 0.21 & 0.44 \\
Herring & 0.02 & 0.02 & 0.03 & 0.13 & 0.16 & 0.03 & 0.07 \\
HouseTwenty & 0.04 & 0.06 & 0.07 & 0.14 & 0.74 & 0.13 & 0.35 \\
InlineSkate & 0.06 & 0.08 & 0.40 & 0.25 & 2.79 & 0.45 & 1.18 \\
InsectEPGRegularTrain & 0.02 & 0.02 & 0.03 & 0.11 & 0.43 & 0.07 & 0.21 \\
InsectEPGSmallTrain & 0.01 & 0.02 & 0.04 & 0.12 & 0.36 & 0.06 & 0.18 \\
InsectWingbeatSound & 0.03 & 0.02 & 0.09 & 0.23 & 1.26 & 0.21 & 0.48 \\
ItalyPowerDemand & 0.01 & 0.01 & 0.01 & 0.11 & 0.07 & 0.02 & 0.06 \\
LargeKitchenAppliances & 0.08 & 0.10 & 0.21 & 0.44 & 1.27 & 0.24 & 0.45 \\
Lightning2 & 0.01 & 0.02 & 0.03 & 0.12 & 0.18 & 0.04 & 0.10 \\
Lightning7 & 0.02 & 0.01 & 0.05 & 0.14 & 0.11 & 0.02 & 0.06 \\
Mallat & 0.03 & 0.04 & 0.15 & 0.17 & 5.47 & 0.94 & 1.77 \\
Meat & 0.01 & 0.01 & 0.04 & 0.11 & 0.13 & 0.03 & 0.06 \\
MedicalImages & 0.02 & 0.01 & 0.04 & 0.25 & 0.27 & 0.06 & 0.14 \\
MiddlePhalanxOutlineAgeGroup & 0.01 & 0.01 & 0.03 & 0.20 & 0.12 & 0.04 & 0.08 \\
MiddlePhalanxOutlineCorrect & 0.01 & 0.01 & 0.02 & 0.29 & 0.19 & 0.05 & 0.12 \\
MiddlePhalanxTW & 0.01 & 0.01 & 0.04 & 0.22 & 0.11 & 0.04 & 0.08 \\
MixedShapes & 0.11 & 0.17 & 0.24 & 0.67 & 6.73 & 1.20 & 2.49 \\
MixedShapesSmallTrain & 0.04 & 0.04 & 0.14 & 0.20 & 5.77 & 1.01 & 2.06 \\
MoteStrain & 0.01 & 0.00 & 0.01 & 0.11 & 0.24 & 0.05 & 0.13 \\
OliveOil & 0.03 & 0.01 & 0.06 & 0.11 & 0.09 & 0.03 & 0.04 \\
OSULeaf & 0.06 & 0.03 & 0.16 & 0.22 & 0.44 & 0.09 & 0.19 \\
PhalangesOutlinesCorrect & 0.04 & 0.02 & 0.03 & 0.97 & 0.56 & 0.19 & 0.37 \\
Phoneme & 0.05 & 0.08 & 0.93 & 0.77 & 4.86 & 0.93 & 1.98 \\
PigAirwayPressure & 0.08 & 0.08 & 1.96 & 0.63 & 1.43 & 0.27 & 0.61 \\
PigArtPressure & 0.07 & 0.08 & 2.12 & 0.60 & 1.43 & 0.27 & 0.54 \\
PigCVP & 0.06 & 0.08 & 1.98 & 0.62 & 1.44 & 0.30 & 0.62 \\
Plane & 0.01 & 0.01 & 0.03 & 0.13 & 0.08 & 0.02 & 0.04 \\
PowerCons & 0.02 & 0.01 & 0.01 & 0.13 & 0.13 & 0.03 & 0.07 \\
ProximalPhalanxOutlineAgeGroup & 0.01 & 0.01 & 0.03 & 0.19 & 0.13 & 0.05 & 0.08 \\
ProximalPhalanxOutlineCorrect & 0.01 & 0.01 & 0.02 & 0.28 & 0.18 & 0.05 & 0.12 \\
ProximalPhalanxTW & 0.01 & 0.01 & 0.03 & 0.20 & 0.13 & 0.03 & 0.08 \\
RefrigerationDevices & 0.05 & 0.09 & 0.08 & 0.44 & 1.26 & 0.27 & 0.59 \\
Rock & 0.05 & 0.05 & 0.35 & 0.14 & 0.46 & 0.10 & 0.19 \\
ScreenType & 0.09 & 0.08 & 0.13 & 0.43 & 1.25 & 0.30 & 0.49 \\
SemgHandGenderCh2 & 0.06 & 0.12 & 0.13 & 0.42 & 3.09 & 0.68 & 1.57 \\
SemgHandMovementCh2 & 0.11 & 0.20 & 0.33 & 0.77 & 3.14 & 0.78 & 1.63 \\
SemgHandSubjectCh2 & 0.14 & 0.19 & 0.29 & 0.72 & 3.13 & 0.79 & 1.62 \\
ShapeletSim & 0.01 & 0.01 & 0.01 & 0.11 & 0.23 & 0.05 & 0.13 \\
ShapesAll & 0.16 & 0.13 & 0.86 & 1.71 & 1.42 & 0.34 & 0.60 \\
SmallKitchenAppliances & 0.05 & 0.08 & 0.10 & 0.40 & 1.25 & 0.29 & 0.46 \\
SmoothSubspace & 0.01 & 0.01 & 0.01 & 0.11 & 0.02 & 0.01 & 0.02 \\
SonyAIBORobotSurface1 & 0.01 & 0.00 & 0.01 & 0.10 & 0.10 & 0.03 & 0.06 \\
SonyAIBORobotSurface2 & 0.01 & 0.00 & 0.01 & 0.10 & 0.15 & 0.04 & 0.09 \\
StarLightCurves & 0.39 & 0.56 & 0.35 & 1.04 & 21.16 & 4.04 & 6.29 \\
Strawberry & 0.04 & 0.03 & 0.04 & 0.34 & 0.55 & 0.14 & 0.25 \\
SwedishLeaf & 0.02 & 0.02 & 0.09 & 0.34 & 0.35 & 0.09 & 0.18 \\
Symbols & 0.01 & 0.01 & 0.05 & 0.11 & 0.90 & 0.20 & 0.34 \\
SyntheticControl & 0.01 & 0.01 & 0.03 & 0.17 & 0.09 & 0.03 & 0.07 \\
ToeSegmentation1 & 0.01 & 0.01 & 0.02 & 0.11 & 0.17 & 0.04 & 0.08 \\
ToeSegmentation2 & 0.01 & 0.01 & 0.02 & 0.11 & 0.13 & 0.03 & 0.07 \\
Trace & 0.01 & 0.01 & 0.04 & 0.13 & 0.13 & 0.04 & 0.08 \\
TwoLeadECG & 0.01 & 0.01 & 0.01 & 0.10 & 0.22 & 0.05 & 0.11 \\
TwoPatterns & 0.05 & 0.03 & 0.05 & 0.69 & 1.46 & 0.33 & 0.84 \\
UMD & 0.01 & 0.01 & 0.02 & 0.11 & 0.07 & 0.02 & 0.04 \\
UWaveGestureLibraryAll & 0.38 & 0.34 & 0.42 & 1.13 & 9.54 & 1.97 & 3.71 \\
UWaveGestureLibraryX & 0.10 & 0.08 & 0.13 & 0.79 & 3.17 & 0.70 & 1.19 \\
UWaveGestureLibraryY & 0.08 & 0.06 & 0.17 & 0.83 & 3.18 & 0.69 & 1.21 \\
UWaveGestureLibraryZ & 0.13 & 0.06 & 0.13 & 0.82 & 3.15 & 0.63 & 1.20 \\
Wafer & 0.03 & 0.03 & 0.03 & 0.66 & 2.44 & 0.48 & 1.18 \\
Wine & 0.01 & 0.01 & 0.02 & 0.11 & 0.07 & 0.02 & 0.04 \\
WordSynonyms & 0.02 & 0.02 & 0.18 & 0.33 & 0.56 & 0.12 & 0.22 \\
Worms & 0.03 & 0.05 & 0.14 & 0.27 & 0.55 & 0.13 & 0.25 \\
WormsTwoClass & 0.03 & 0.05 & 0.06 & 0.24 & 0.55 & 0.13 & 0.25 \\
Yoga & 0.03 & 0.04 & 0.06 & 0.31 & 3.10 & 0.55 & 1.18 \\
\hline
\textbf{Sum} & 5.88 & 5.75 & 18.38 & 53.16 & 158.77 & 34.56 & 73.47 \\
\hline
\end{longtable}

\begin{table}[!h]
\centering
\caption{Computation time (in minutes) for multivariate datasets.} 
\begin{tabular}{c|cccc}
\hline
Dataset & RidgeCV & RandomForest & LogRegCV & LDA \\
\hline
DuckDuckGeese & 0.18 & 0.18 & 0.18 & 0.16 \\
PEMS-SF & 0.87 & 0.99 & 0.84 & 0.58 \\
FaceDetection & 0.57 & 0.61 & 0.65 & 0.57 \\
MotorImagery & 0.47 & 0.50 & 0.47 & 0.52 \\
Heartbeat & 0.65 & 0.72 & 0.67 & 0.72 \\
FingerMovements & 0.58 & 0.49 & 0.59 & 0.56 \\
NATOPS & 0.73 & 0.78 & 0.74 & 0.76 \\
PhonemeSpectra & 0.05 & 0.09 & 0.05 & 0.04 \\
HandMovementDirection & 0.54 & 0.47 & 0.58 & 0.49 \\
ArticularyWordRecognition & 0.87 & 0.98 & 0.97 & 0.97 \\
SelfRegulationSCP2 & 0.43 & 0.47 & 0.44 & 0.46 \\
BasicMotions & 0.63 & 0.73 & 0.63 & 0.35 \\
Cricket & 0.82 & 0.89 & 0.92 & 0.93 \\
EigenWorms & 0.50 & 0.52 & 0.53 & 0.44 \\
LSST & 0.30 & 0.51 & 0.25 & 0.26 \\
RacketSports & 0.72 & 0.85 & 0.76 & 0.55 \\
SelfRegulationSCP1 & 0.73 & 0.82 & 0.77 & 0.73 \\
ERing & 0.95 & 0.95 & 0.96 & 0.88 \\
StandWalkJump & 0.60 & 0.47 & 0.53 & 0.20 \\
Epilepsy & 0.31 & 0.47 & 0.33 & 0.33 \\
EthanolConcentration & 0.48 & 0.43 & 0.65 & 0.81 \\
Handwriting & 0.17 & 0.20 & 0.24 & 0.15 \\
UWaveGestureLibrary & 0.67 & 0.84 & 0.78 & 0.53 \\
AtrialFibrillation & 0.33 & 0.33 & 0.40 & 0.20 \\
Libras & 0.52 & 0.74 & 0.63 & 0.51 \\
\hline
\end{tabular}
    \label{tab:dd_mtsc}
\end{table}
\end{document}